\documentclass[conference]{IEEEtran}
\usepackage{cite}
\usepackage{graphicx}
\usepackage{color}
\usepackage{booktabs}
\usepackage{multirow}
\usepackage[cmex10]{amsmath}
\usepackage{siunitx}
\usepackage[tight,footnotesize]{subfigure}
\usepackage{dblfloatfix} 
\usepackage{float}
\usepackage{caption}
\usepackage{breqn}
\usepackage{graphicx}
\usepackage{lscape}
\usepackage[numbers,sort&compress]{natbib} 

\begin{document}
\title{2D-Motion Detection using SNNs with Graphene-Insulator-Graphene Memristive Synapses}

\author{Shubham Pande,
         Karthi Srinivasan,
         Suresh Balanethiram,
         Bhaswar Chakrabarti,
		  Anjan Chakravorty~\IEEEmembership{Member,~IEEE}
		}

\maketitle
\begin{abstract} 
The event-driven nature of spiking neural networks makes them biologically plausible and more energy-efficient than artificial neural networks. In this work, we demonstrate motion detection of an object in a two-dimensional visual field. The network architecture presented here is biologically plausible and uses CMOS analog leaky integrate-and-fire neurons and ultra-low power multi-layer RRAM synapses. Detailed transistor-level SPICE simulations show that the proposed structure can accurately and reliably detect complex motions of an object in a two-dimensional visual field.    
\end{abstract}

\IEEEpeerreviewmaketitle
\section{Introduction}
\IEEEPARstart{I}{n} the field of computer vision, motion detection has been traditionally performed using machine learning algorithms on hardware that are themselves based on the von Neumann architecture. While modern-day microprocessors face the infamous memory bottleneck issue, these approaches often turn out to be power-hungry \cite{mem_wall}. On the other hand, spiking neural networks (SNNs) emulate the dynamics of biological neural networks in electronic circuits and deliver low-power, inherently parallel computation. Event-driven computation thus becomes an alternative in realizing energy-efficient hardware for numerous cognitive tasks \cite{snn1}. The emergence of novel non-volatile memory technologies like resistive random access memories (RRAMs) \cite{NVM_intro}, phase change memories \cite{PCM_intro}, and spin transfer torque RAMs \cite{spinotronics_intro} has further accelerated the development of SNNs. 
\begin{figure}[h!]
\centering
\includegraphics[width=0.5\textwidth]{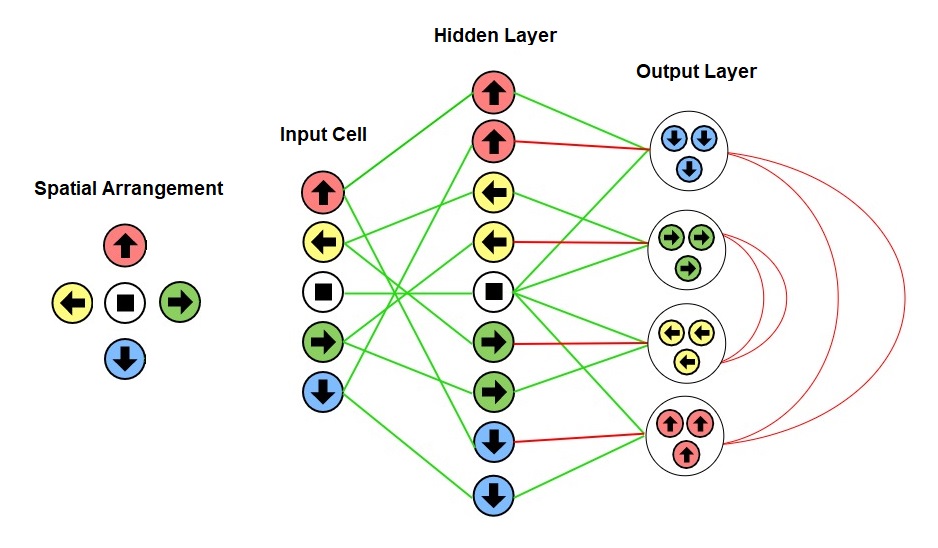}
\caption{(Left) The spatial arrangement of the input layer of a single unit cell. (Right) Unit cell architecture  used for 2-D motion detection, showing the neurons in the input, hidden, and output layers. LIF neurons represented by circles are interconnected by RRAM-based  synapses shown by lines.}
\label{network_architecture}
\end{figure}
Several techniques have been proposed in the past for motion detection using SNNs \cite{Yang_2018_MD1,fnbot_MD2,AER_MD,Senn_W_2002_MD_STDP,Feidler_1997_MD_Hebbian,buchs2002spike}. In \cite{Yang_2018_MD1}, a novel temporal coding scheme is used to encode the real-time motion into a series of spikes and a gradient descent learning algorithm is utilized to train the two-layered SNN. An autonomous neuromorphic agent that could perform obstacle-avoidance and target acquisition was demonstrated using the CMOS-based ROLLS neuromorphic processor in \cite{fnbot_MD2}. Various studies such as probabilistic feed-forward synapses \cite{Senn_W_2002_MD_STDP}, rate-based Hebbian learning \cite{Feidler_1997_MD_Hebbian} and spike timing-dependent plasticity \cite{buchs2002spike} have been proposed to realize directional selectivity and consequently motion detection. On the other hand, considering biological neural systems' high efficacy and accuracy, 
authors in \cite{Dalgaty_et_al} have shown insect-inspired elementary motion detection. 
It is envisaged that the ability to mimic an insect's navigation system would have a significant impact on the field of robotics, compact visual surveillance, etc., due to the limited computational capacity of their on-board processor.

In our effort to develop biologically inspired, low power motion detection framework, we improve and extend the delay and correlate network presented in \cite{Dalgaty_et_al} to detect complex motions. Our proposed SNN based network uses CMOS-based analog leaky integrate-and-fire (LIF) neurons and ultra-low-power, forming-free RRAMs as synapses. We present an end-to-end SPICE implementation of the complete network using an experimentally validated physics-based Verilog-A model of the synapse and transistor-level implementations for the neurons. We improve the network's directional resolution by incorporating lateral inhibition. Also we show that the network can detect a wide range of input signal frequencies by adding an appropriate number of output neurons per direction. 

This paper is organized as follows. Section II details the network architecture. Section III discusses the physics-based compact model of the GIG synapse, LIF neuron circuit, and peripheral circuitry used for stimulus generation. Section IV presents the simulation results followed by a conclusion in section V.

\section{Network Architecture}
To detect the motion of an object moving along a particular direction, a delay and correlate network was presented in \cite{Dalgaty_et_al}. Basically, the idea is to introduce a temporal delay between two spatially adjacent inputs followed by a downstream mechanism for detecting the spatio-temporal correlations. Here, we modify the network presented in \cite{Dalgaty_et_al} to network shown in Fig.~\ref{network_architecture}. Spatial arrangement of the input layer of a unit cell is shown in Fig.~\ref{network_architecture} (Left). The neuron at the center (white circle) is called a neutral neuron as it is not pointing in any of the cardinal directions.
Neurons in the input layer are connected to the output layer neurons through a hidden layer as shown in Fig.~\ref{network_architecture}. We introduce lateral inhibition (shown by red lines) between Up-Down and Left-Right output neurons to emphasize the fact that an object can't move in Up-Down or Left-Right direction at the same time. This modification reduces the spiking activity for output neurons that are not meant to fire for given excitation and improve the network accuracy. Additionally, in order to make the network capable of detecting objects moving with a wide range of frequencies, we use N (instead of using a single) output neurons per direction. The value of N and time constants of each output neuron can be adjusted to meet the application-specific requirements.

\begin{figure}[t!]
\centering
\includegraphics[width=0.31\textwidth]{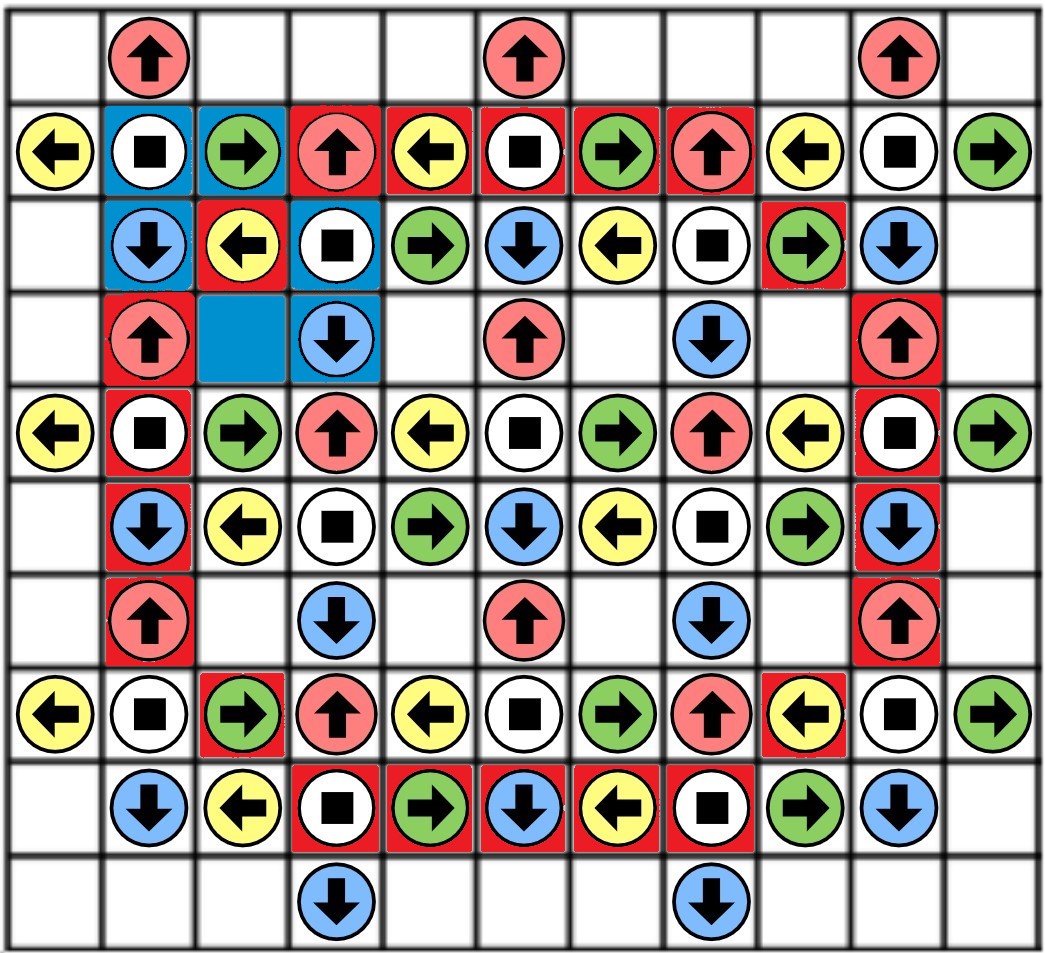}
\caption{The tessellation pattern is consisting of 15 unit cells, arranged in a staggered manner for a 10$\times$11 visual field. Red denotes the path of motion of the center of the 3$\times$3 square stimulus, while the stimulus itself is shown in blue. }
\label{visual_field}
\end{figure}

\begin{figure}[t!]
\centering
\includegraphics[width=0.5\textwidth]{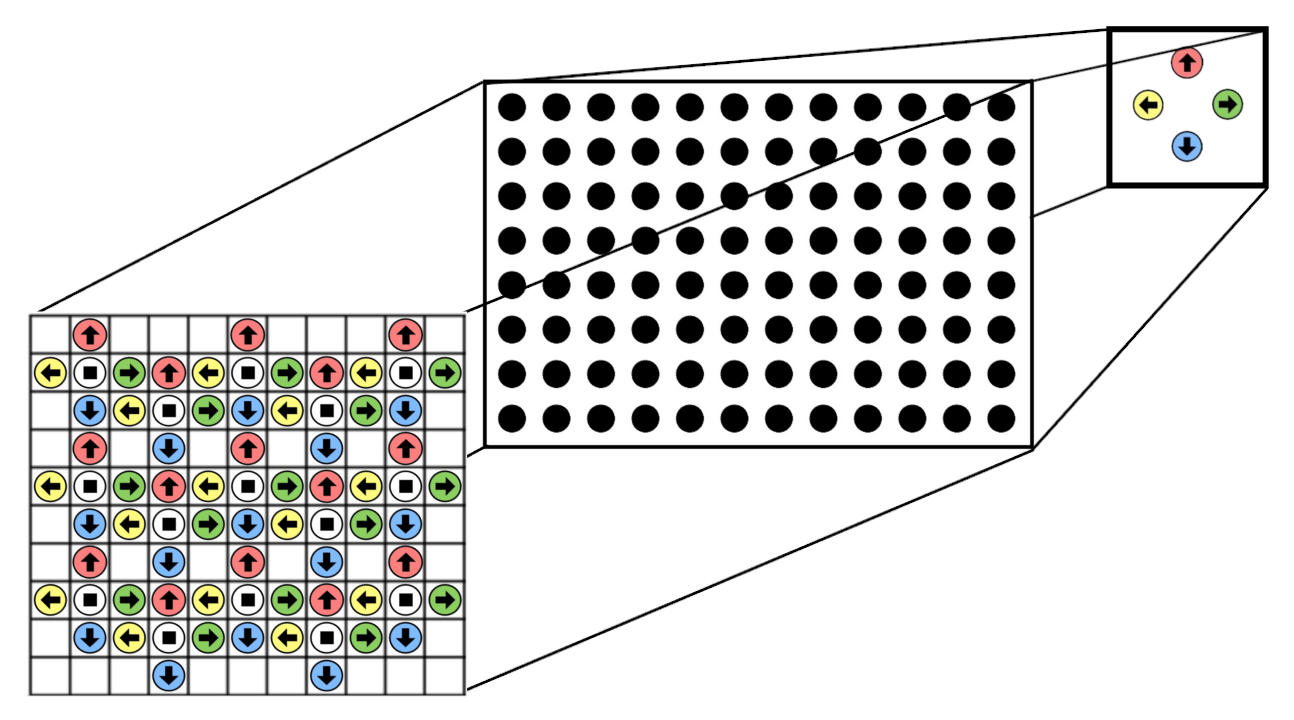}
\caption{Structure of the implemented spiking neural network to detect the motion of an object across a visual field. Here neurons (shown by circles) in one layer are connected to neurons in other layers through synapses (not shown to avoid cluttering).}
\label{network_topology}
\end{figure}

The operation of the network shown in Fig.~\ref{network_architecture} is explained below. Input layer neurons denote the spatial activity of the object in the visual field and encode this activity by introducing a temporal delay between activity in spatially adjacent regions. This information from the input layer is passed on to the hidden layer through synapses. The adjustment of time constants of hidden layer neurons coupled with information received from the input layer generates a unique spatio-temporal pattern, encoding the spatial activity at the input layer. In this case, the time constant of the neutral neuron is made smaller than the other four neurons. Finally, the spatio-temporal pattern generated at the hidden layer is transmitted to the output layer. The incoming spikes to the output layer may have an excitatory (green lines) or inhibitory (red lines) effect on the output layer. The output layer neurons are designed with threshold voltages and time constants such that they will fire if they receive two consecutive excitatory spikes within a short time interval. However, if an inhibitory spike is received shortly before the first spike, then the membrane voltage is reduced sufficiently and the excitatory pair of spikes does not cause an output spike.
\begin{figure}[h!]
\centering
\includegraphics[width=0.5\textwidth, height=0.35\textwidth]{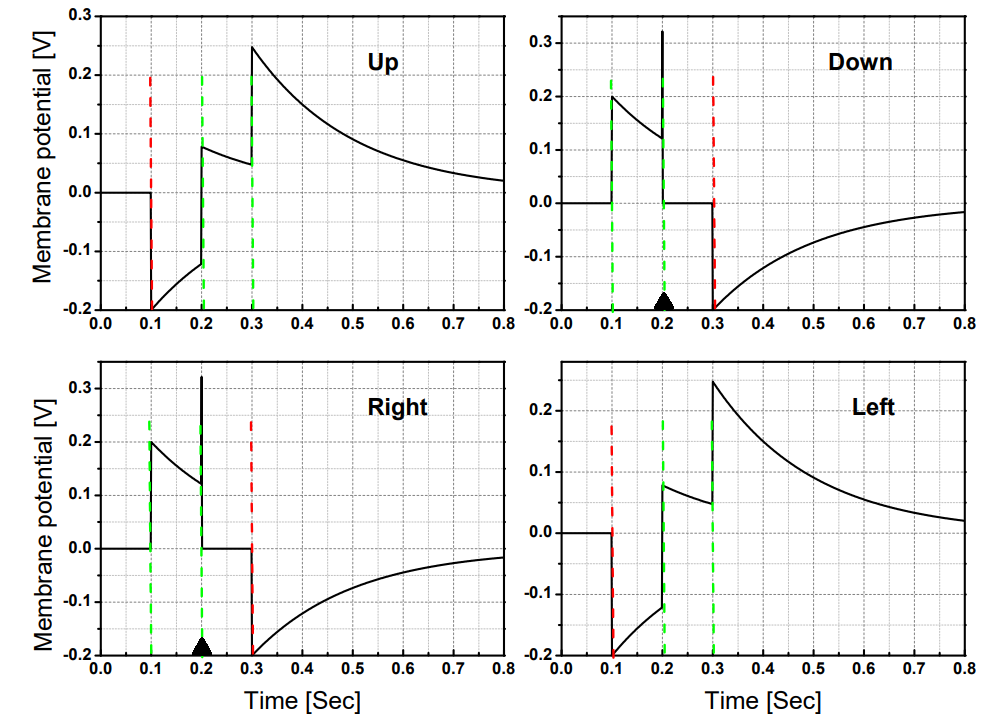}
\caption{The evolution of membrane potentials of the output neurons in the case where the stimulus is moving diagonally from Left-Up to Right-Down. Vertical dotted lines denote input spikes (green for excitatory, red for inhibitory). Triangle markers denote output spikes.}
\label{sample_spikes}
\end{figure}
Fig.~\ref{sample_spikes} illustrates the working of the unit cell when the object moves in a specific direction across an input cell, say, diagonally from Left-Up to Right-Down. The input neurons get excited in the order: Left and Up; Center; Right and Down. The temporal delays in the activity at spatially adjacent location are encoded at the input layer and coupled with time constant adjustments of the hidden layer neurons, which results in a spike pattern at the output neuron as follows: Down and Right receive two excitatory spikes followed by one inhibitory. Up and Left receive one inhibitory spike followed by two excitatory. Hence, the Right and Down neurons spike for this input while the other two do not. The input cell shown in Fig.~\ref{network_architecture} is used to tessellate the 10$\times$11 visual field as shown in Fig.~\ref{visual_field}. Each of the 15 input cells has its own patch of hidden layer neurons which connects to a common output layer as shown in Fig.~\ref{network_topology}. The network uses a total of 75 input neurons, 135 hidden neurons, and 4N output neurons, where N was varied from 1 to 5 for performance comparison. A total of 315 synapses are required to realize the connections between the neurons.
\section{SPICE Implementation}
\subsection{Synapse Model}
Resistive random-access memory technology has emerged in the last decade as a promising candidate for applications in non-volatile storage as well as for neuromorphic computation \cite{intro_IEDM,intro_ISSCC_2011,intro_VLSIT_2009,intro_VLSIT_2012,intro_nature,intro_TED}. "Forming-free" operation with ultra-low operating current has been previously observed in graphene-insulator-graphene (GIG) RRAM devices with chemical vapor deposited (CVD) graphene electrodes \cite{Bhaswar_et_al_G-I-G}. The switching material for the GIG device is a three-layer stack of TiO$_x$, Al$_2$O$_3$, and TiO$_{2}$. The incorporation of CVD-grown graphene as electrodes significantly reduces the operating current compared to conventional electrodes such as TiN. The ultra-low ON and OFF state currents (180 nA and $\sim$100pA, respectively) and the ON-state non-linearity make these devices highly desirable for high-density storage and compute-in-memory applications. For a detailed discussion of GIG device fabrication and characterization, see \cite{Bhaswar_et_al_G-I-G}. A physics-based SPICE-compatible compact model to make this device suitable for large-scale circuits and systems design can be found here \cite{GIG_model}. The electrical equivalent schematic of the RRAM used for the simulation is shown in Fig.~\ref{LIF} (top). The model framework assumes the existence of a pre-formed filament(s) in the top TiO$_2$ and the bottom TiO$_x$ layers because of the forming-free nature of the device. The total series resistance offered by graphene electrode and TiO$_2$ layer is modeled as a lumped resistor of value R$_{gr}$ + R$_{TiO_{2}}$. The resistive switching behavior in this device is attributed to the formation and rupture of a conductive filament in the Al$_{2}$O$_{3}$ layer, and it is modeled as a series connection of variable resistor (R$_{Al_{2}O_{3}}$) and dependent current source (I$_{gap}$). For more details about the model formulation, we direct readers to \cite{GIG_model}. 

\subsection{Analog LIF neuron}
\begin{figure}[t!]
\centering
\includegraphics[width=0.5\textwidth]{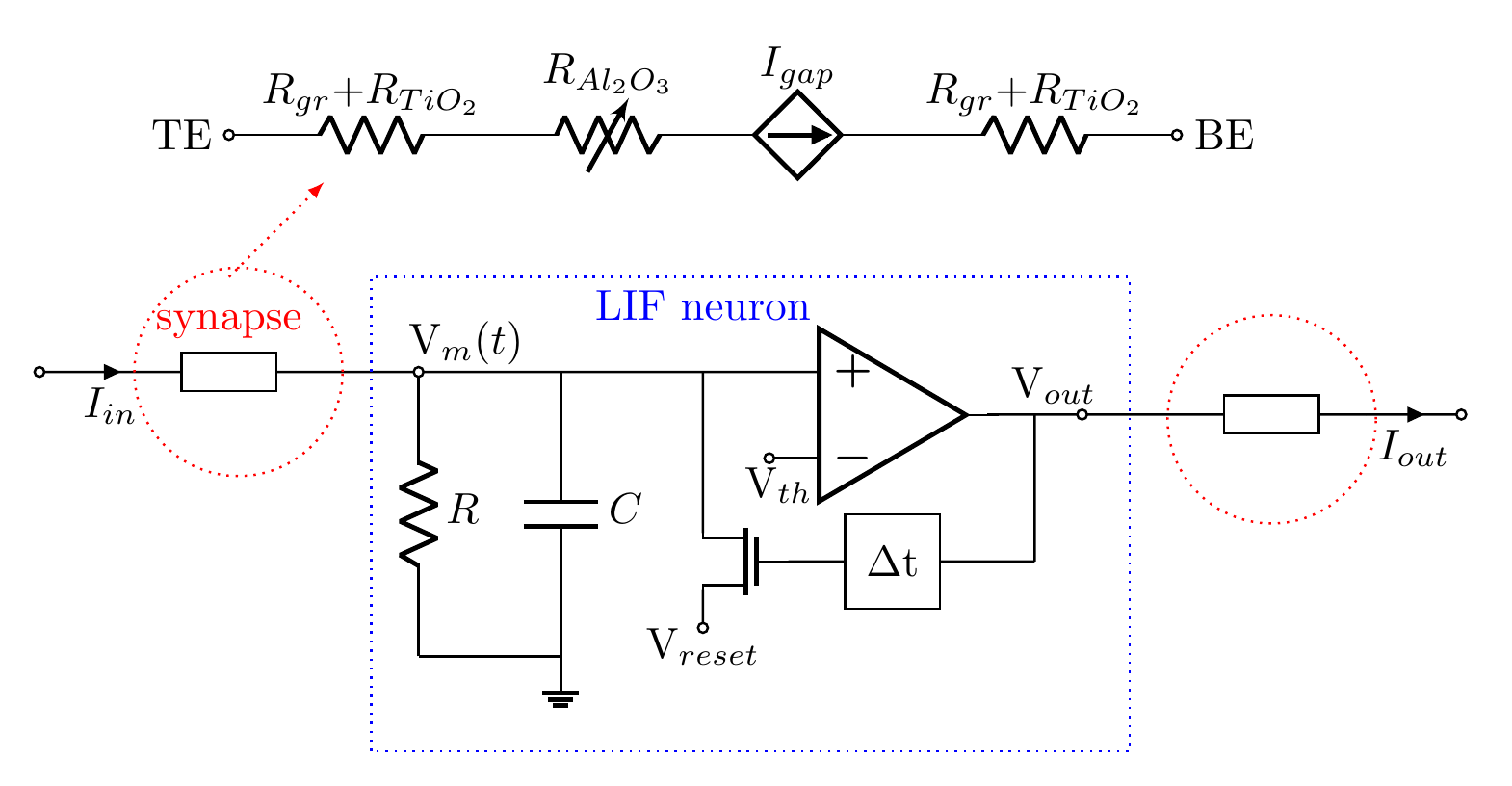}
\caption{(Top) Electrical equivalent of the RRAM. (Bottom) Schematic of a Leaky Integrate-and-Fire (LIF) neuron and the incoming and outgoing RRAM synapses.}
\label{LIF}
\end{figure}

The LIF neuron model is one of the most widely used models owing to its computational efficiency and biological plausibility \cite{analog_LIF_neuron}. The LIF model consists of a parallel combination of a resistor (R) and an integrating capacitor (C), as shown in Fig.~\ref{LIF}. The currents I$_{in}$ coming from the synapses, connected at the input terminal of the LIF neuron, charge up the capacitor. Each spiking neuron is characterized by an internal state variable, the membrane potential V$_{m}$(t). When this membrane potential exceeds a fixed threshold (V$_{th}$), a spike is sent out, and the capacitor is discharged to a reset potential (V$_{reset}$) using the voltage-controlled NMOS switch. A comparator and a delay block are used to generate the constant-width output spikes. The neuron block is implemented using TSMC's 180nm CMOS standard cell library.

\subsection{Stimulus Generation}
The input spikes denoting the motion of the object are generated using a peripheral circuit that mimics a dynamic vision sensor (DVS). Each pixel in a DVS is sensitive to relative temporal contrast in the luminous intensity, and upon detection of a sufficiently significant change, it sends out a spike \cite{DVS_sensor}. To emulate this behavior, we treat each event as a vector $e = (x, y, t)$, where $x$ and $y$ define the pixel location in the visual field, and $t$ is the time of the event. The trajectory of the moving object is modeled using the appropriate mathematical function and it is used to excite the input layer.
\begin{figure*}[t!]
\centering
\includegraphics[scale=0.27]{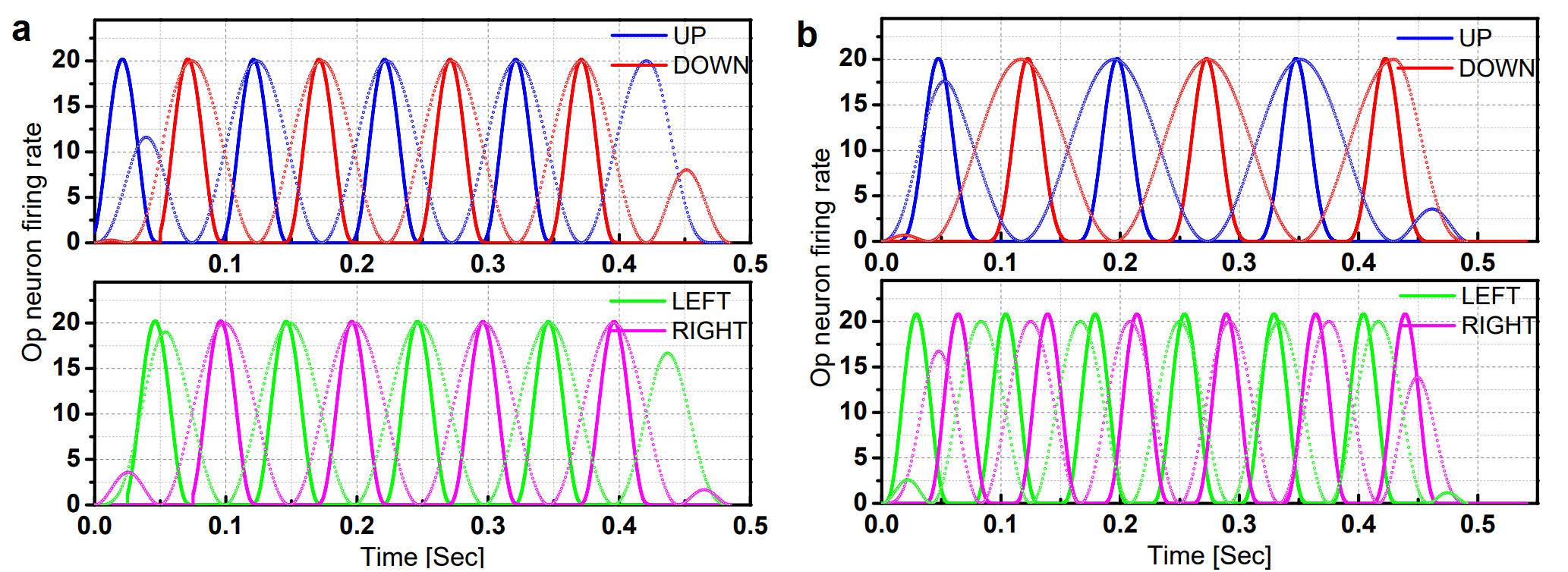}
\caption{Ideal (dotted lines) and SPICE simulated (solid lines) firing rates of the output neuron for an object moving in (a) a circular path and (b) an eight-shaped path.}
\label{Circle_infinity}
\end{figure*}
\section{Results}
A SPICE simulation of the network shown in Fig.~\ref{network_topology} is done using Cadence Virtuoso \cite{cadence}. The object is assumed to be a square block of 3$\times$3 pixels, shown in blue in Fig.~\ref{visual_field}. The red color denotes the circular path followed by the object's center.
Whenever the object is at a specific location, all neurons that the object covers produce a spike that propagates through the deeper layers of the network as shown in Fig.~\ref{network_topology}. We expect a significant increase in the firing activity of the output neuron
when the object moves across the visual field along its corresponding direction. The output firing rates for circular and eight-shaped trajectories are shown in Fig.~\ref{Circle_infinity}. 
For a circular input path, the four output neurons produce sinusoidal firing rates, with relative phase lags of 90$^{\circ}$ between them, as shown in Fig.~\ref{Circle_infinity}(a). This can be understood by considering the UP-DOWN and LEFT-RIGHT neuron pairs as two orthogonal basis elements encoding the direction of motion. The stimulus travels rightwards from 0 to T/2, leftwards from T/2 to T, upwards from 3T/4 to T/4, and downwards from T/4 to 3T/4 where T is the time period. The firing rates of the corresponding output neuron reach their maxima at the midpoint of the time intervals mentioned above. Similarly, for an eight-shaped input path, the firing rate of the UP-DOWN neurons is half of that of the LEFT-RIGHT neurons. Here, the stimulus makes two oscillations along the horizontal direction for every oscillation along the vertical direction as shown in Fig.~\ref{Circle_infinity}(b).\\
The instantaneous firing rate ($f^o$) is calculated by filtering the spike trains using a low-pass filter as 
\begin{equation}
    f^o(t)=h(t)*S(t)
\end{equation}
where * denotes the convolution operation. $S(t)$ refers to the spike train (a sequence of impulses at the locations of the spikes) and $h(t)=\lambda(e^{-t/\tau_1}-e^{-t/\tau_2})u(t)$ refers to the impulse response of the aforementioned filter with $\tau_1$ and $\tau_2$ as constants that are set to the mean and twice the mean of the output neuron time constants. $\lambda$ is a normalization constant and $u(t)$ is the unit step function.

\begin{figure}[h!]
\centering
\includegraphics[scale=0.48]{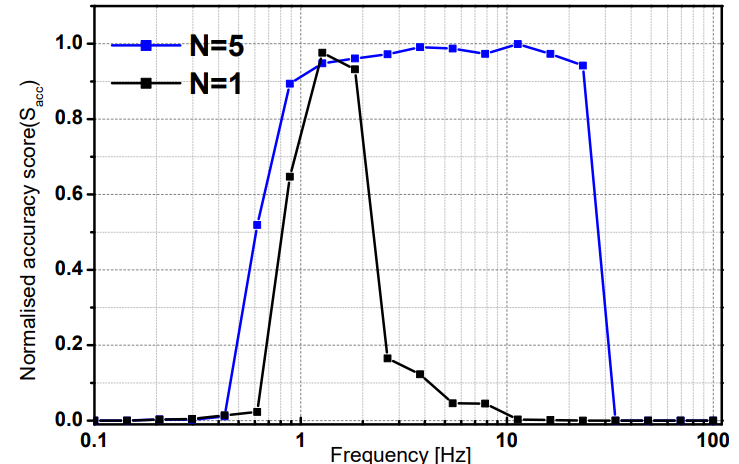}
\caption{Normalized accuracy score vs. input motion frequency. Using a larger number of output neurons with a range of time constants (5 ms-500 ms vs. 500 ms) allows for reliable motion detection over a wider range of frequencies.}.
\label{rms}
\end{figure}

The ideal spiking rates ($f^o_{ideal}$) corresponding to four output neurons are obtained from the parameterization ($x(t),y(t)$) of the path for considered trajectories (circular and eight shaped) as
\begin{equation}
    f^o_{i,ideal}(t)=\frac{f_{max}}{2}\left|\frac{\Dot{p}(t)}{\Dot{p}_{max}}+1\right|
\end{equation}
where $i$ takes on four values depending on the output channel (UP, DOWN, LEFT, RIGHT). For UP and DOWN (LEFT and RIGHT), the directional velocity $\dot{p}(t)$ and its maximum value $\dot{p}_{max}$ are replaced by $\dot{y}(t)$ and $\dot{y}_{max}$ ($\dot{x}(t)$ and $\dot{x}_{max}$), respectively. Here $f_{max}$ is the maximum firing rate of the output neurons. The ideal firing rates reach their maximum when the object moves at maximum velocity along the preferred direction and go to zero when the object moves against the preferred direction. In case of no movement either along or against, they settle to half the maximum rate. The firing rate obtained from SPICE simulations (solid lines) is showing a good level of match with the ideal firing rate (dotted lines) as shown in Fig.~\ref{Circle_infinity}.\\
In order to make the network sensitive to a wide range of moving object frequencies, we use N output neurons (instead of one neuron) per direction. The network with five output neurons per direction, with time constants being logarithmically distributed from 5 ms to 500 ms, performs better than the network with one output neuron per direction, with a 500 ms time constant. The accuracy score ($S_{acc}$) is shown in Fig.~\ref{rms} as a function of the rotation frequency for the circular input case. The accuracy score is calculated as
\begin{equation}
    S_{acc}=\frac{1}{4}\sum_i \left(1-\frac{|f^o_{i,ideal}(t)-f^o_{i,measured}(t)|^2}{|f^o_{i,ideal}(t)|^2}\right).
\end{equation}
A network designed with an appropriate value of N will be able to identify the required range of moving object frequency.

\section{Conclusion}
We have demonstrated end-to-end SPICE simulation of SNN based motion detection framework. We design our network using an experimentally validated physics based synapse model and transistor-level implementation of LIF neurons. We demonstrated improvement in network performance by including lateral inhibition and pool of N output neurons per direction. The proposed network can accurately encode the object's direction and velocity of movement over a significant range of velocities. A notably similar area where the same network can be applied is event-based tactile sensing where the stimulus is localized and moves at relatively low velocities across the sensory field. The network can help to extract velocity information directly from the input for using in higher-level applications like handwriting recognition. Tactile sensing data-sets such as the ST-MNIST digits data-set \cite{See2020STMNISTT} provide a good target case for this network architecture beyond its original intended application. Other future work also includes the processing of Braille character data collected using tactile neuromorphic sensors.
\ifCLASSOPTIONcaptionsoff
  \newpage
\fi
\bibliographystyle{unsrtnat}

\bibliography{rram_ref}
\end{document}